\documentclass[english,conference]{article}
\usepackage[T1]{fontenc}
\usepackage[latin9]{inputenc}
\usepackage{geometry}
\usepackage{fancyhdr}
\usepackage{float}
\usepackage{amsmath}
\usepackage{amssymb}
\usepackage{graphicx}
\usepackage{setspace}
\usepackage[numbers]{natbib}
\pdfoutput=1
\makeatletter

\floatstyle{ruled}
\newfloat{algorithm}{tbp}{loa}
\providecommand{\algorithmname}{Algorithm}
\floatname{algorithm}{\protect\algorithmname}

\numberwithin{equation}{section}
\numberwithin{figure}{section}

\usepackage{algpseudocode} 
\usepackage{graphicx}

\@ifundefined{showcaptionsetup}{}{%
 \PassOptionsToPackage{caption=false}{subfig}}
\usepackage{subfig}
\makeatother

\usepackage{babel}
\begin{document}

\title{\date{}Learning to Fingerprint the Latent Structure in Question
Articulation}

\author{Kumar Mrityunjay$^{+}$ and Guntur Ravindra$^{++}$\\
Technology Excellence Group$^{+}$, Senior Member of the ACM$^{++}$\\
Email: mjay.cse@gmail.com$^{+}$, gravindra@acm.org$^{++}$\\
Talentica Software, Pune, India}
\maketitle
\begin{abstract}
Machine understanding of questions is tightly related to recognition
of articulation in the context of the computational capabilities of
an underlying processing algorithm. In this paper a mathematical model
to capture and distinguish the latent structure in the articulation
of questions is presented. We propose an objective-driven approach
to represent this latent structure and show that such an approach
is beneficial when examples of complementary objectives are not available.
We show that the latent structure can be represented as a system that
maximizes a cost function related to the underlying objective. Further,
we show that the optimization formulation can be approximated to building
a memory of patterns represented as a trained neural auto-encoder.
Experimental evaluation using many clusters of questions, each related
to an objective, shows 80\% recognition accuracy and negligible false
positive across these clusters of questions. We then extend the same
memory to a related task where the goal is to iteratively refine a
dataset of questions based on the latent articulation. We also demonstrate
a refinement scheme called $K$-fingerprints, that achieves nearly
100\% recognition with negligible false positive across the different
clusters of questions. 
\end{abstract}
\thispagestyle{fancy}
\fancyhf{}
\fancyhead[C]{ACCEPTED FOR PRESENTATION AT the 17th IEEE INTERNATIONAL CONFERENCE ON MACHINE LEARNING AND APPLICATIONS (ICMLA2018)}

\section{Introduction}

In practical applications, algorithms that power natural language
interfaces designed to answer questions have to grapple with various
types of user inputs and articulation styles. Inputs could be grammatically
incorrect, they could be in a natural language that the algorithms
are not trained for, and they may be too complex for the current algorithm
to interpret. Despite challenging input variations, these interfaces
are expected to respond back even if they are unable to interpret
the input. Interpretation is always in the context of an underlying
algorithm that performs the task of answer generation. The complexity
of this underlying algorithm can vary depending on answer data as
well as question interpretation capabilities in relation to the answer
data. We refer to these aspects of answerability as an ``objective''.
We argue that question answerability is always in the context of the
objective. As a result, we need a system that can find a mapping between
the latent structure of articulation and the algorithm that processes
the question for a underlying knowledge-base. 

\pagebreak{}

\subsection{Motivation}

\fancyhf{}The challenging input conditions that question-answering
systems have to grapple with, motivates us to explore how pattern
recognition techniques may be used to detect if input questions have
latent patterns that are friendly to an objective. The natural language
interface then uses this pattern recognition system to detect if the
input question matches the objective, and if it does, the objective
is triggered. In practical applications we only have data that can
be friendly to an objective, but the complementary dataset is a vast
universe of possibilities for which we do not have complete information.
Hence the system needs to account for the complementary information
in the pattern recognition system without having explicit information
of this complementary set. In this paper we address this issue by
first formulating a mathematical model for the same, and then approximating
the model's implementation to that of a neural auto-encoder. We then
show how the same model can be used to iteratively learn and re-partition
a data set of questions based on a latent structure in the articulation
style.

The rest of this paper is organized as follows: in section-\ref{sec:Literature-Review}
we present a literature review covering techniques for question analysis
in some applications. In section-\ref{sec:Recognizing-inputs-based}
a theoretical formulation for the proposed pattern recognition system
is presented. An iterative learning process called $K$-fingerprints
is detailed in section-\ref{sec:Iterative-re-partitioning} followed
by experimental evaluation in section-\ref{sec:Experiments-and-Results},
and a conclusion in section-\ref{sec:Conclusion-and-Future}.

\section{Literature Review\label{sec:Literature-Review}}

Analysis of question structure and its impact on performance of answering
systems has been well studied in the context of natural language knowledge
repositories \cite{open-domain-qa}. The general trend in question
classification has been to use the nature of the answer classes ,
and identification of a factual intent (wh-type) \cite{function-based-question-classification}.
A large class of question answering techniques tie the question domain
to the knowledge domain and its representation. In recent years there
has been growing interest in analyzing question structure in the context
of answerability by humans. In \cite{language-use} the authors present
a supervised learning approach to predict if a question shall be answered
or not. They propose use of a trained SVM classifier where multiple
features such as length of the question, n-grams matching in a known
corpus, POS diversity, LDA topic diversity, question editing activities,
and characteristics of topic hierarchy are used to determine answerability.
In another research \cite{question-question}, which also used a SVM
classifier, a mix of boolean features, counting-based features and
scoring-based features were used to classify questions in terms of
answerability. The authors also analyzed answerability in terms of
classifier performance on question types based on elicited response
types such as advise, fact, opinion. In a related work \cite{not-answered-questions},
detection of questions that can go unanswered using heuristic features
such as question length, asker history, polite words, and question
subjectivity was presented. In \cite{will-q-answered}, question answerability
was modeled using a combination of title tokens, body tokens, sentiment,
user data, and time. In summary, the goals of popular research in
question answerability is to be able to model the relationship between
question structure and its answerability so as to predict the answerability
itself.

Contrary to popular literature, techniques presented in this paper
have a slightly different goal, although related to answerability
of questions. An underlying natural language processing algorithm
designed to function deterministically for a specific objective works
well when the latent structure in question articulation matches what
the algorithm has been designed for. Hence here we are interested
in representing and detecting the latent structure in articulation
of a question. 

One possible approach to implement such a detector is to build a set
of Bloomfilters \cite{bloomfilter} each representing one objective.
Every question is now passed to all the Bloomfilters, and the filter
that recognizes the pattern in the question identifies the question
with the corresponding objective. However, one would have to still
handle false positives as multiple Bloomfilters could mark the question
as detected. Further, the choice of number of bits in the Bloomfilter
depends on expected false positive which in-turn can de determined
if all the possible variations in question articulation is available
apriori. In cases where we have small application specific data this
requirement cannot be satisfied. Yet another possibility is to use
locality sensitive hashing \cite{lsh, plsh} where questions with
similar latent patterns can be hashed to the same hash bucket. Detection
would then entail hashing an input question's latent patterns to a
bucket, and selecting the label assigned to the bucket as the objective.
However, in order to achieve high recall and zero false positive one
would have to use large number of hash functions and account for unseen
data. Further, hashing based schemes fundamentally do not account
for any non-linearities in the latent structure of articulation when
such a structure can be numerically transformed.

\section{Recognizing inputs based on an objective \label{sec:Recognizing-inputs-based}}

In the context of question style recognition, we define an objective
$\mathcal{O}$ as something that can successfully determine if the
question that is being asked can be answered by the underlying system.
Recognition of an objective depends on the ability to algorithmically
understand the articulation of a question. The underlying recognition
system also has data requirements. For example if the user asks a
question related to Soccer, the system can generate answers only if
data related to Soccer is present in the data store. As another example,
if the data is stored in a relational database, the system can generate
an answer via a database query only if it is capable of figuring out
the correct database columns from the words in the question. Yet another
example is a system that employs an algorithm that can convert questions
starting with ``how many'' into a SQL query, but the same algorithm
could generate the wrong query for any other type of questions. Hence
defining the objective $\mathcal{O}$ is a key starting point.

Objectives that are algorithm and data dependent are very difficult
to model as a computational entity. Hence we propose a workflow that
involves (1) representing questions as feature vectors (2) clustering
questions based on feature vectors (3) human evaluation of sampled
questions from clusters (4) selection of question clusters that match
$\mathcal{O}$ (5) modeling latent patterns in articulation of questions
as a fingerprint (6) using fingerprints to detect if a question matches
$\mathcal{O}$. In order to implement these steps we need a formal
approach to defining the problem.

\subsection{Problem formulation}

Let $Q=\{q_{1},q_{2},.....,q_{n}\}$ be a set of questions. We define
a transformation $\mathcal{T}(q_{i})\rightarrow s_{i}$ where $s_{i}$
is a mathematical representation of $q_{i}$. The set $S=\{s_{1},s_{2},...,s_{n}\}$
is the corresponding mathematical representation of $Q$. We define
an open set $S^{'}\subseteq S$ where members of $S^{'}$ meet the
objective $\mathcal{O}$. We are interested in a function $f(*)$
such that

\begin{equation}
\begin{array}{c}
f(s_{i})\rightarrow\mathbb{R}_{i}^{d}\forall s_{i}\in S^{'}\\
f^{-1}(\mathbb{R}_{i}^{d})\in S^{'}
\end{array}\label{eq:core_function}
\end{equation}

Such a function can transform any question into a $d$ dimensional
feature vector with a constraint that an inverse transformation would
result in one of the elements in $S^{'}$. An identity function could
have satisfied such a relationship, but this is not very useful as
questions unrelated to $\mathcal{O}$ would also be wrongly identified
as being related to $\mathcal{O}$. Hence we need an additional constraint
on $f(*)$ defined as 

\begin{equation}
f^{-1}(\mathbb{R}_{k}^{d})\notin S^{'}\forall s_{k}\in S-S^{'}\label{eq:constraint_on_core_function}
\end{equation}

In (\ref{eq:constraint_on_core_function}), we say that any question
from outside the set $S^{'}$ would result in an inverse that does
not belong to $S^{'}$ and therefore does not match $\mathcal{O}$.
In most practical applications where $\mathcal{O}$ is known, one
can determine $S^{'}$ easily. But the complementary set $\widetilde{S^{'}}=S-S^{'}$
is unknown and difficult to find. Hence we need to find the $f(*)$
without the knowledge of $\widetilde{S^{'}}$. However, before we
can discover $f(*)$ we need to define $\mathcal{T}$.

\subsection{Defining the transformation function $\mathcal{T}$ \label{subsec:Defining-the-transformation}}

We need to define the transformation $\mathcal{T}(q_{i})$ so that
a question in natural language can be converted to a form suitable
for computational manipulation. The algorithmic steps for this transformation
are shown in algo-(\ref{Algo:transform_question_to_vector}). We believe
that this approach is most suitable when one wants to identify questions
that are relevant for a given $\mathcal{O}$. 

\begin{figure*}
\includegraphics[scale=0.5]{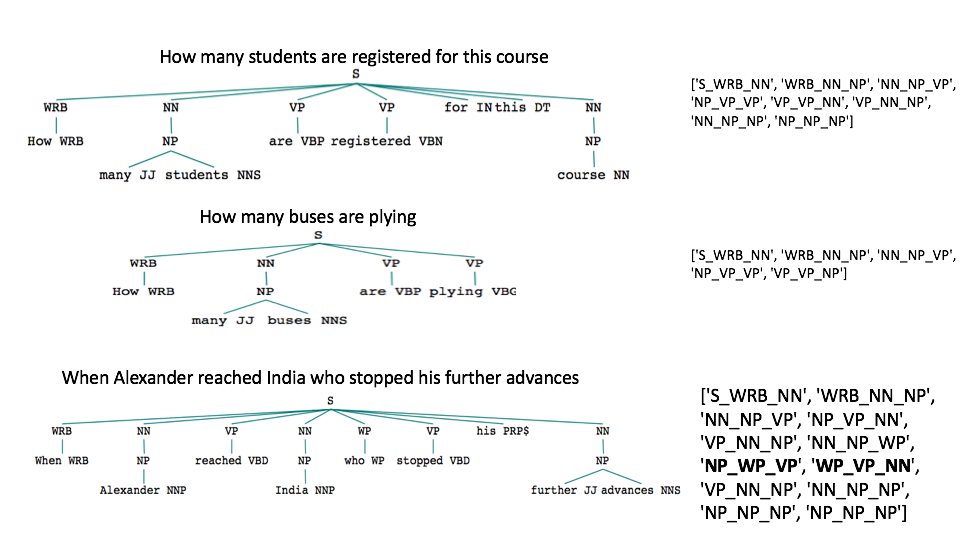}

\caption{Parse trees for different questions and their traversal sequences}
\label{fig:parse-tree}
\end{figure*}

One can model a question as a sequence of words from a finite vocabulary.
However, specific words and word orderings are used while expressing
information as a question. When we generate a parse tree for a question
$(PARSE(q_{i}))$, the tree is a representation of the underlying
structure in expressing information as a question. However, there
can be different trees for different questions, and questions of similar
types have similar trees. The notion of similarity in questions is
dependent on $\mathcal{O}$. Example of parse trees is shown in figure-\ref{fig:parse-tree}.
In order to be able to compare the structure of trees, we use a string-matching
approach. We traverse the nodes $(TRAVERSE(T_{i}))$ of the parse
tree to generate a sequence of node labels. This sequence is represented
as partial orders with a known rule ($SEQUENCES(B_{i})$). For example,
one could convert a sequence into non-overlapping segments of fixed
length by skipping even indices. In addition one could create additional
segments by skipping odd indices. For example, in figure-\ref{fig:parse-tree}
we have three questions that have different parse trees. Two of the
questions have a similar parse tree. If we perform a breadth-first
search and aggregate trigrams of parse nodes, we get sets of trigrams
that are similar across all the trees. However that last tree has
additional trigrams that are different from those for the first two
trees. This difference can be captured as a difference in articulation. 

Using this information a binary feature vector is created such that
existence of a particular symbol/sequence is represented by $1$ and
$0$ otherwise. Algo-\ref{Algo:transform_question_to_vector} describes
the various steps involved in converting a set of questions into the
mathematical representation referred to in (\ref{eq:core_function}).
On line-\ref{line:weight} in algo-\ref{Algo:transform_question_to_vector},
the variable $\tau(x)$ is a weight. This weight can be an importance
value derived relative to all the other $x\in D$. It could also be
an indicator variable that takes a value $1$. In this paper we use
the indicator variable version.

\begin{algorithm}
\begin{algorithmic}[1]

\Require$Q$ is a non-empty set

\State$D\leftarrow\{\}$

\ForAll {$q_{i}\in Q$}

\State $T_{i}\leftarrow PARSE(q_{i})$

\State $B_{i}\leftarrow TRAVERSE(T_{i})$

\State $C_{i}\leftarrow SEQUENCES(B_{i})$

\State$D\leftarrow D\cup C_{i}$

\EndFor

\Ensure$D$ is a non-empty and a unique set of elements

\Ensure D is an ordered set

\Ensure$d=\left|D\right|$

\Ensure$INDEX(x,D)$ is the position of $x$ in the ordered set $D$

\ForAll{$q_{i}\in Q$ }

\State $\mathbb{R}^{d}=0$

\ForAll{$x\in C_{i}$}

\State $\mathbb{R}\left[INDEX(x,D)\right]=\tau(x)$\label{line:weight}

\EndFor

\State $\mathcal{T}(q_{i})\rightarrow s_{i}=\mathbb{R}^{d}$

\EndFor

\end{algorithmic}

\caption{Transforming questions into vectors}
\label{Algo:transform_question_to_vector}

\end{algorithm}

\subsection{Discovering  $f\left(*\right)$}

The asymmetric function $f(*)$ in (\ref{eq:core_function}) is a
one-to-one transformation where as its inverse (\ref{eq:constraint_on_core_function})
is a one-to-many transformation. If we find a function whose inverse
is also one-to-one transformation, then an identity function would
have sufficed. But an identity cannot account for the additional constraint
(\ref{eq:constraint_on_core_function}). Interestingly, we need to
determine this function without the explicit knowledge of the complementary
set $\widetilde{S^{'}}$. In order to make the function discovery
feasible we employ two computational strategies. We assume a universe
of functions $\mathcal{F}$, of which $f^{*}\in\mathcal{F}$ is the
optimal function satisfying (\ref{eq:core_function}) and the constraint
(\ref{eq:constraint_on_core_function}). Secondly, to satisfy the
constraint (\ref{eq:constraint_on_core_function}) without the knowledge
of the complementary set $\widetilde{S^{'}}$, we relax the one-to-one
transformation to a many-to-one transformation of a reduced dimensionality.
This means $f(s_{i})$ results in a vector of dimensionality much
lower than that of $s_{i}$. The inverse continues to be a one-to-many
transformation where we constrain the transformation to any element
in set $S^{'}$. As a result of this relaxation we can then express
the function discovery as a mathematical expression shown in (\ref{eq:error_defined_for_auto_encoder-1}) 

\begin{equation}
\arg\max_{f\in\mathcal{F}}\frac{1}{\left|S^{'}\right|}\sum_{s_{i}\in S^{'}}\max_{s_{k}\in S^{'}}\left(SIM\left[f^{-1}\left(f\left(s_{i}\right)\right),s_{k}\right]\right)\label{eq:error_defined_for_auto_encoder-1}
\end{equation}

where, $SIM$ is a measure of similarity.

\subsection{Practical implementation as a fingerprinting system \label{subsec:Practical-implementation}}

The optimization function in (\ref{eq:error_defined_for_auto_encoder-1})
can be implemented as an encoder-decoder formulation where $f$ represents
the encoder and $f^{-1}$ is the decoder. The encoder-decoder combine
needs to be jointly optimized in order to implement (\ref{eq:error_defined_for_auto_encoder-1}).
The transformation $\mathcal{T}$ described earlier converts each
question into a vector of weights of a fixed dimensionality. Hence
one of the frameworks that can jointly optimize the encoder-decoder
combine is a neural auto-encoder with multiple hidden layers. We use
an eight layer auto-encoder with Tanh as the activation function for
each hidden layer, whereas one hidden layer has a ReLU activation
\cite{relu}. The output layer has linear activation. The chosen Adam
optimizer \cite{adam} runs for 300 epochs with a learning rate of
0.001 and\textbf{ }mean square error as the loss function. We then
define the similarity function $SIM$ mentioned in (\ref{eq:error_defined_for_auto_encoder-1})
as shown in algo-\ref{Algo: similarity-function}. This function measures
the detection accuracy on a scale of $0$ to $1$. A perfect reconstruction
of the input in the output layer of a trained auto-encoder would qualify
for a detection accuracy of $100\%$ and algo-\ref{Algo: similarity-function}
guarantees this. 

In the remainder of the paper we use the terms fingerprinting system
and auto-encoder interchangeably as the fingerprinting technique is
executed by a trained auto-encoder.

\begin{algorithm}
\begin{algorithmic}[1]

\Ensure$\tau=1$

\State $X$ is the input to the function $f$

\State $Y=f^{-1}\left(f\left(X\right)\right)$

\State$ILARGEST([],\ n)$ takes vector as input and returns index
of $n$ largest values

\State $A\leftarrow$indices $i$ of $X$ where $X(i)=\tau$ 

\State $B\leftarrow ILARGEST(Y,|A|)$

\State $SIM\leftarrow\frac{\left|A\cap B\right|}{\left|A\cup B\right|}$

\end{algorithmic}

\caption{Defining the similarity function $SIM()$}
\label{Algo: similarity-function}
\end{algorithm}

\section{$K$-fingerprints \label{sec:Iterative-re-partitioning}}

The fingerprinting system described in the previous section can be
used to iteratively refine clusters by learning new latent patterns.
Further, it can be used to refine clusters of similarly articulated
questions by eliminating members that do not confirm to a pattern.
We refer to this technique as $K$-fingerprints as we use a set of
$K$ trained auto-encoders $AE=\{ae_{1}^{t},ae_{2}^{t},...,ae_{K}^{t}\}$.
Steps implemented in this algorithm are summarized in algo-\ref{algo:iterative-training}.

To begin with, a large dataset of questions is first clustered into
$K$ clusters $\hat{C}=\{\hat{C_{1}},\hat{C_{2}},...,\hat{C_{K}}\}$
using an algorithm like K-means. Owing to inaccuracy in the clustering
algorithm, it is possible that some of the clusters do not have questions
with a common latent pattern. At time step $t=0$, a random sample
of questions is selected from each cluster to create a base set of
$K$ clusters $C=\{C_{1}^{0},C_{2}^{0},....,C_{K}^{0}\}$ whose latent
patterns are to be learned by the $K$ auto-encoders (one per cluster).
We also maintain a set of cluster cardinalities $M=\{m_{1}^{0},$$m_{2}^{0},...,m_{K}^{0}\}$
and these are updated in every iteration. 

\subsection{Iterative re-training with filtering\label{subsec:Iterative-re-training-with}}

At $t=0$, $C_{i}^{t=0}$ is used to train the auto-encoder $ae_{i}^{t=0}$
resulting in baseline auto-encoders. The baseline auto-encoders are
iteratively retrained over multiple time steps until a convergence
criterion is met. Details of iterative training is given in algo-\ref{algo:iterative-training}.
The iterative training process involves two major phases, namely $DETECT$,
and $TRAIN$. In the $DETECT$ phase at step $t$, questions from
$\hat{C_{i}}$ are passed as input to $ae_{i}^{t-1}$. In time step
$t-1$ the auto-encoder $ae_{i}^{t-1}$ would have been trained using
questions from $C_{i}^{t-1}$ of cardinality $m_{i}^{t-1}$. At time-step
$t$, each auto-encoder $ae_{i}^{t-1}$ is made to detect all possible
questions in $\hat{C}_{i}$. Say $ae_{i}^{t-1}$ detects a set of
$C_{i}^{t}\subseteq\hat{C}_{i}$ questions. If the cardinality of
$C_{i}^{t}$ is not the same as $m_{i}^{t-1}$ then the $i^{th}$
auto-encoder survives the current round. If there is no change, then
the $i^{th}$ auto-encoder is eliminated from participation in subsequent
rounds and is added to another set $F$. All the auto-encoders that
have survived the next round, now move on to the $TRAIN$ phase. In
the $TRAIN$ phase, auto-encoder $ae_{i}^{t-1}$ is trained using
questions from $C_{i}^{t}$ to result in $ae_{i}^{t}$ and $M$ is
updated to $M=\{m_{1}^{t},m_{2}^{t},....,m_{K}^{t}\}$ where $m_{i}^{t}=\left|C_{i}^{t}\right|$.
This process is repeated until $F$ ends up with $K$ trained auto-encoders
or we have looped through the process enough number of times. After
the iteration loop ends, auto-encoders that are not part of $F$ now
become part of $F$ provided $m_{i}^{t}>0$.

\begin{algorithm}
\begin{algorithmic}[1]

\Require $\hat{C}=\{\hat{C_{1}},\hat{C_{2}},...,\hat{C_{K}}\}$ a
set of $K$ clusters of questions 

\Require$C=\{C_{1}^{0},C_{2}^{0},....,C_{K}^{0}\}$ a set of $K$
clusters where $C_{i}^{0}\subset\hat{C}_{i}$ 

\Require$AE=\{ae_{1}^{0},ae_{2}^{0},....,ae_{K}^{0}\}$ a set of
$K$ baseline auto-encoders trained with $C$

\Require$M=\{m_{1}^{0},m_{2}^{0},....,m_{K}^{0}\}$ a set of integers
showing how many questions were detected by the baseline auto-encoders

\Require $F$ is the final set of trained auto-encoders

\State$G\leftarrow AE$

\State Initialize iteration round $t\leftarrow1$

\While {$G$ is not empty}

\State$G^{t}=\{\phi\}$ , $C^{t}=\{\phi\}$, $M^{t}=\{\phi\}$

\For{Each auto-encoder $AE_{i}^{t-1}$in $G$}

\State$C_{i}^{t}\leftarrow\mathbf{DETECT(AE_{i}^{t-1},\hat{C}_{i})}$ 

\State \textbf{Continue} if $C_{i}^{t}$ is empty

\If{ items detected $\left|C_{i}^{t}\right|\ne$ $m_{i}^{t-1}$}

\State Include $AE_{i}^{t-1}$ in $G^{t}$

\State Include $C_{i}^{t}$ in $C^{t}$

\State Include $\left|C_{i}^{t}\right|$ in $M^{t}$

\Else

\State Include $AE_{i}^{t-1}$ into $F$

\EndIf

\EndFor

\State$M\leftarrow M^{t}$, $C\leftarrow C^{t}$, $AE=\{\phi\}$

\For{$AE_{i}\in G^{t}$}

\State$AE_{i}^{t}\leftarrow\mathbf{TRAIN(AE_{i},C_{i}^{t})}$

\State Include updated $AE_{i}^{t}$ into $AE$ 

\EndFor

\State $G\leftarrow AE$ (Updated list of auto-encoders)

\State \textbf{Loop counter to exit}

\State$t\leftarrow t+1$ (Next Iteration)

\EndWhile

\State $F\leftarrow F\cup C$ (Final list of trained auto-encoders)

\State Use each $AE_{i}$ in $F$ and detect classes of questions
in $\hat{C}_{i}$ using the $DETECT()$ function

\end{algorithmic}

\caption{Training K-fingerprints}
\label{algo:iterative-training}

\end{algorithm}

\section{Experiments and Results \label{sec:Experiments-and-Results}}

Approaches described in the previous sections are evaluated under
two broad criteria. On one hand we want the fingerprint system to
detect questions with articulation that matches its memory. At the
same time, we want each fingerprint system to fail on detecting questions
used to train the other fingerprint systems. The second criterion
is harder to meet as there are $K-1$ complementary memories for every
fingerprint system and there can be some questions that have an articulation
which could match with multiple memories. In other words there can
be confusion among multiple fingerprint systems. In applications such
as chat-bots or translation systems we require the fingerprint system
to have 0\% false positives and a very high true positive rate in
detecting the articulation type. In these applications a false positive
would result in wrong interpretation or an out-of-context answer where
as a low true positive would result in frequent no-response situations.
Unlike typical machine learning applications, here we are not interested
in unseen data (test data). We are interested in seeing if a pattern
match system learned the known set of patterns well. Keeping this
type of scenario in mind we evaluate the fingerprinting system in
terms of true positive rate and false positive rate. 

\subsection{Data set}

To evaluate the approaches described in the previous sections, SQuAD
data set \cite{sqa-dataset} (dev-data set version-1.1) was used as
a source of questions with various types of articulations. There were
around 10k questions spanning 48 different topics. Questions were
transformed into feature vectors using the transformation process
described in section-\ref{subsec:Defining-the-transformation}. For
the tree-traversal, we used a breadth-first-search over the parse
tree and converted sequence of parse labels into trigrams and 4-grams.
We used an overlap of 2 and 3 for trigrams and 4-grams respectively.

\begin{figure}
\includegraphics[scale=0.2]{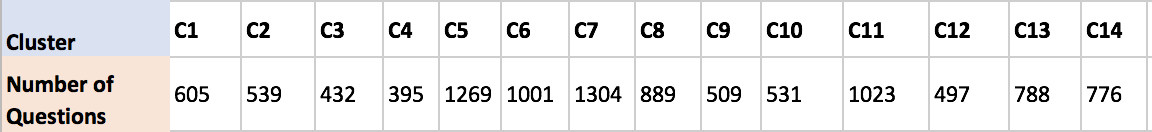}

\caption{Number of questions in each cluster}
\label{fig:num-questions-per-cluster}

\end{figure}

This resulted in a dimensionality of 192 unique symbols (algo-\ref{Algo:transform_question_to_vector}).
Each question in turn, was transformed into a 192 dimensional feature
vector where each appearing symbol was given a weight of $\tau=1$.
The questions in the feature vector form were clustered into 14 clusters
using K-means++ \cite{kmeans++}. The choice of 14 was based on inspecting
the density of points in bins of a K-d tree. The goal of clustering
is to help a human evaluator to identify consistent sets of clusters
by randomly sampling and viewing questions in each cluster. The expectation
is that questions in each cluster correspond to a specific type of
articulation, and therefore different machine interpretability. Figure-\ref{fig:num-questions-per-cluster}
shows the number of questions in each of the 14 clusters. We observe
that some clusters have many more questions than others but there
are at least 350 questions per cluster.

\subsection{Baseline evaluation\label{subsec:Baseline-evaluation}}

\begin{table}
\includegraphics[scale=0.25]{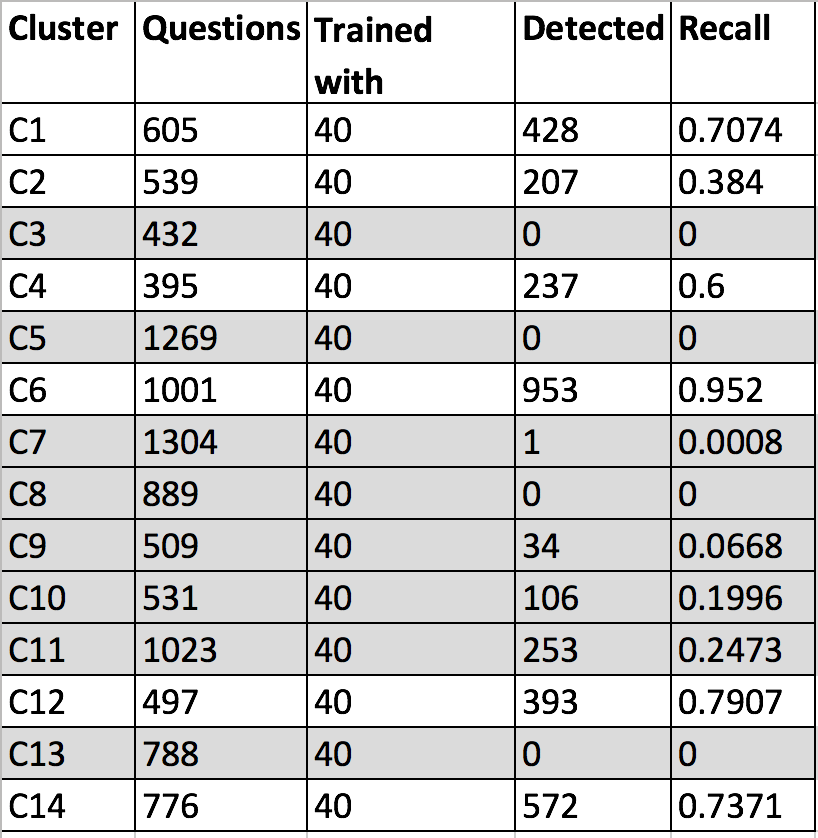}

\caption{Accuracy of baseline fingerprint systems}
\label{tab:results-baseline-ae}

\end{table}

From each of the 14 clusters 40 questions were selected per cluster
at random. These questions were used to train 14 fingerprint systems
(one per cluster) to get the baseline fingerprint systems. Note that
the dimension of the input vector varies for each fingerprint system,
and it depends on the total number of unique symbols in each cluster.
Each baseline fingerprint system was then made to detect all the questions
in the corresponding cluster. The goal was to observe how well the
fingerprint system trained with a small set of questions can detect
all the remaining questions in a cluster. Table-\ref{tab:results-baseline-ae}
shows the number of questions detected by each baseline fingerprint
system for the corresponding cluster. We observe that many fingerprint
systems have a poor recall. This indicates that the corresponding
clusters did not have questions with similar articulation. If the
articulation would have been the same, we expect the feature vectors
to be the same, and therefore a higher recall. The issue could be
with poor clustering, and is resolved using $K$-fingerprints as explained
later. 

After we exclude clusters with low recall we have 6 clusters namely
$\{C_{1},C_{2},C_{4},C_{6},C_{12},C_{14}\}$. The questions in each
of these selected clusters were used to train a corresponding fingerprint
system with parameters as described in section-\ref{subsec:Practical-implementation}.
Note that each fingerprint system learns to detect questions corresponding
to the specific objective as represented by the articulations in the
corresponding cluster. The trained fingerprint systems are now used
to detect questions from the 6 clusters. We are interested in true
positive for fingerprint system $ae_{i}$ when questions in $C_{i}$
are used as an input and the false positive when questions from other
than $C_{i}$ are used as an input to $ae_{i}$.

\begin{figure}
\subfloat[$SIM$>=0.5]{\includegraphics[scale=0.18]{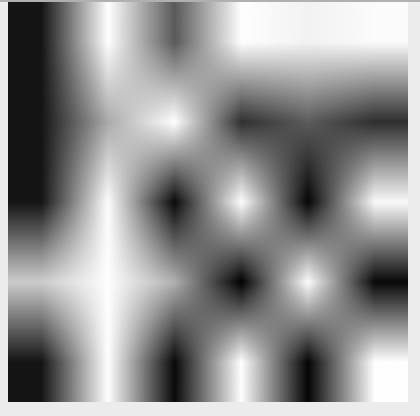}

}\subfloat[$SIM$>=0.7]{\includegraphics[scale=0.18]{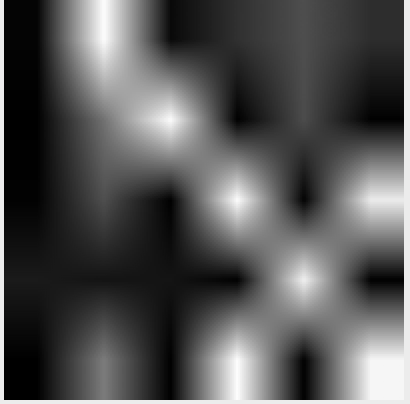}

}\subfloat[$SIM$>=0.9]{\includegraphics[scale=0.18]{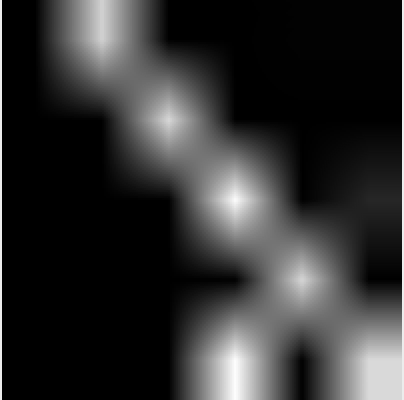}

}

\caption{Confusion matrix for different thresholds viewed as a heat-map}
\label{fig-confusion-heat-map}
\end{figure}

\begin{figure}
\subfloat[Recognition accuracy and confusion matrix for selected clusters]{\includegraphics[scale=0.4]{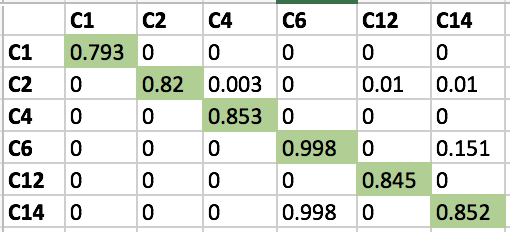} 

\label{fig:confusion-90}

}

\subfloat[Fingerprints showing that $C6$ and $C14$ are very similar with $C14$
being a superset of $C6$]{\includegraphics[scale=0.25]{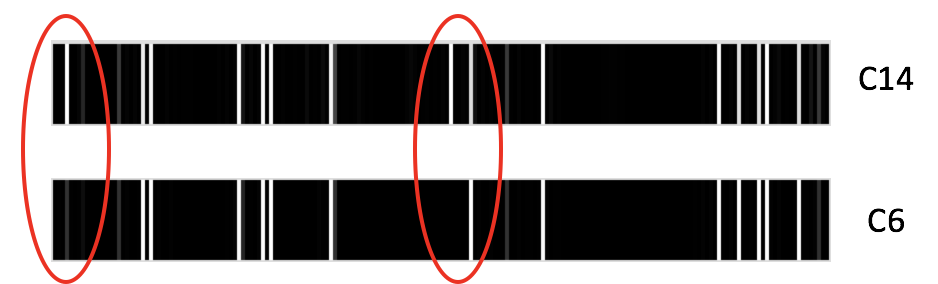}

\label{fig:c6_vs_c14}

}

\caption{Confusion matrix for the 6 clusters and cluster similarity for $C14$
and $C6$ }
\end{figure}

\begin{table*}
\includegraphics[scale=0.18]{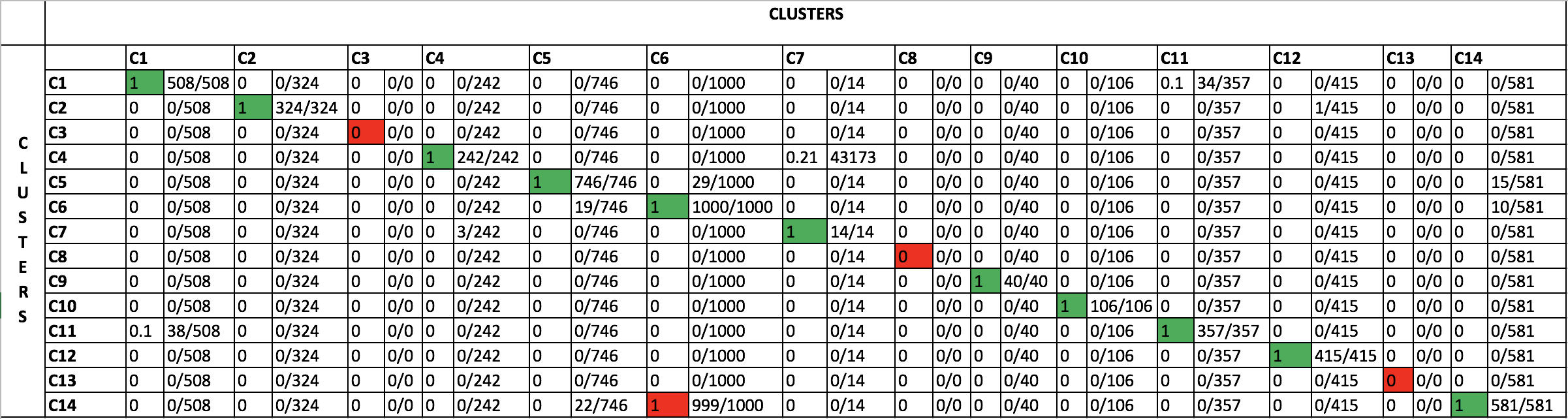}

\caption{Recognition accuracy, confusion matrix and cluster membership}
\label{tab:k-fingerprints-results}

\end{table*}

\subsection{Evaluation}

In order to evaluate the accuracy of the fingerprinting system, a
6x6 confusion matrix was generated corresponding to the selected clusters.
Each $(i,j)$ cell in the confusion matrix refers to the percentage
of questions detected as relevant to the $i^{th}$ cluster by the
$i^{th}$ fingerprinting system, when questions from the $j^{th}$
objective/cluster were given as an input. Ideally one would like this
value to be $0$ when $i\ne j$ and $1$ when $i=j$. Figure-\ref{fig-confusion-heat-map}
shows the confusion matrix viewed as a heat-map when different thresholds
were used on the $SIM()$ function described in algo-\ref{Algo: similarity-function}.
Note how the accuracy improves as the threshold increases from 0.5
to 0.9. Ideally we would like to get a bright line along the principal
diagonal when the confusion is zero. We choose the threshold as 0.9
as it imposes hard performance limits in terms of confusion as well
as true positives. Usually when the threshold is high the confusion
reduces but the true positive rate also reduces. In the present case
we see good true positive rate as well as very low confusion (false
positive across clusters).

Figure\textendash \ref{fig:confusion-90} shows the confusion matrix
when the threshold on $SIM()$ is 0.9. We observe that all the fingerprints
performed well in terms of confusion, except the one corresponding
to $C14$. Here, $C14$ was confused with $C6$. In order to understand
why $C14$ was confused with $C6$ but not vice versa, we visually
plot the average of the input feature vector as an image as shown
in \ref{fig:c6_vs_c14}. We observe that the feature vector is very
similar between $C6$ and $C14$, with $C14$ having two additional
symbols ($C14$ is a superset of $C6$). This in turn implies that
the clustering algorithm failed to group questions corresponding to
$C6$ and $C14$ into a single cluster. However, the fingerprinting
system detected the fact that $C6$ is similar to $C14$ and not vice
versa there by justifying the clustering algorithm's behavior.

The performance of fingerprinting in terms of confusion matrix is
very good, and the recall of most fingerprint systems is around 80\%.
A system of this form is very useful in use-cases where false positives
cannot be tolerated where as false negatives can still be tolerated.
Some typical use-cases are like chat-bots, where it is fine if the
bot refuses to answer the question due to the verdict of the fingerprint
system (false negative), but it is hilarious when an unrelated answer
is generated in response to the question (false positive). The fingerprint
system can avoid such a scenario. 

\subsection{K-fingerprints}

As detailed in the previous sections, $K$-fingerprints is an approach
to iteratively train and partition data so as to improve fingerprinting
accuracy. In table-\ref{tab:results-baseline-ae} we observed that
many clusters had a very poor recall, main reason being poor clustering
accuracy. We use $K$-fingerprints to learn to refine clusters after
the $K$ auto-encoders are initially trained with a randomly selected
set of 40 questions per cluster. During the iterative process each
$ae_{i}^{t-1}$ eliminates questions from the $i^{th}$ cluster if
the fingerprint for these questions do not match with the memory of
$ae_{i}^{t-1}$. By refining the set iteratively, $ae_{i}^{t-1}$
also learns new symbols and this appears as a change in input layer
size, where retraining would result in a new memory for $ae_{i}^{t}$.
When questions in a cluster are eliminated in a particular round,
they may be included in another round of training and the repetitive
$TRAIN$ and $DETECT$ steps described in algo-\ref{algo:iterative-training}
ensures this process.

In the present case we have 14 auto-encoders that iteratively refine
the initial 14 clusters of questions. Table-\ref{tab:k-fingerprints-results}
summarizes multiple observations after $K$-fingerprints have been
trained. The table shows the true positive recognition rate highlighted
in green color. We observe that the true positive is nearly 100\%
in all the 14 cases and this is a marked improvement from the previous
case where we dropped a few clusters due to clustering inconsistency.
What $K$-fingerprints did was to eliminate some questions from each
cluster so that it learns a consistent pattern from among the questions
that remained in the cluster. We can observe that the number of questions
in some of the clusters has actually reduced. For example $C1$ and
$C11$ initially had 605 and 1023 questions respectively (table-\ref{tab:results-baseline-ae}).
However, in table-\ref{tab:k-fingerprints-results} we see that the
number of questions has reduced to 508 and 357 respectively. In addition,
the detection accuracy is 100\% for $C1$ and $C11$ where as cluster
$C11$ was earlier dropped out of contention in the initial experiment
due to poor clustering accuracy. In case of $C11$, $K$-fingerprints
eliminated a large set of questions in order to create a consistent
cluster for which the true positive rate is now 100\%. 

Other interesting case is with respect to $C7$ and $C9$. Here $K$-fingerprints
reduced the cluster sizes by a significant number in order to improve
the true positive (better cluster consistency) without impacting the
false positives. Clusters $C3$ and $C8$ could not be improved from
the previous experiment where they were eliminated. They continue
to stay eliminated after $K$-fingerprints. However, $C5$ is interesting
as there was 0\% match in the previous experiment where as after $K$-fingerprints,
the cluster size was reduced from 1269 questions to 746 questions
resulting in a 100\% recognition accuracy (true positive) and 0\%
confusion with questions from other clusters (false positive).

\section{Conclusion and Future Work \label{sec:Conclusion-and-Future}}

In this paper the notion of objective-driven question style detection
was presented as a mathematical model. We argued that style of articulation
of questions can be captured in a pattern recognition system which
we refer to as fingerprinting. We showed how the fingerprinting system
can be implemented using auto-encoders, and explained why an auto-encoder
is a suitable architecture to represent the mathematical model. Algorithms
to convert groups of questions with similar articulation into a mathematical
model of representation were described. Applications in the domains
of chat bots and translation systems were cited and performance evaluation
conducted using a dataset of over 10K questions. 

The idea of fingerprinting was extended to what we referred to as
$K$-fingerprints, where we showed how an iterative process can be
used to refine clusters of questions in order to generate a fingerprint
that can capture consistency among cluster members. We showed that
the recognition accuracy is nearly 100\% for each fingerprint with
almost 0\% error in recognizing fingerprints of unrelated clusters.

As an extension to the idea of $K$-fingerprints, one can now implement
an iterative clustering scheme. Instead of eliminating questions that
do not have a common articulation, we could extend $K$-fingerprints
to move questions among clusters thereby creating newer clusters.

One can observe that once the questions were transformed using the
transformation function, rest of the steps are outside the realm of
natural language and text processing. This motivates us to explore
the possibility of applying the algorithms described to other types
of data such as images. We believe that the mathematics would remain
the same but the computational method for function discovery would
change. It would have to account for the spatial nature of information
in the images. Adding convolutional layers in combination with dense
layers in the auto-encoder would be beneficial. One needs to explore
though.


\begin{thebibliography}{10}
\providecommand{\url}[1]{#1}
\csname url@samestyle\endcsname
\providecommand{\newblock}{\relax}
\providecommand{\bibinfo}[2]{#2}
\providecommand{\BIBentrySTDinterwordspacing}{\spaceskip=0pt\relax}
\providecommand{\BIBentryALTinterwordstretchfactor}{4}
\providecommand{\BIBentryALTinterwordspacing}{\spaceskip=\fontdimen2\font plus
\BIBentryALTinterwordstretchfactor\fontdimen3\font minus
  \fontdimen4\font\relax}
\providecommand{\BIBforeignlanguage}[2]{{%
\expandafter\ifx\csname l@#1\endcsname\relax
\typeout{** WARNING: IEEEtran.bst: No hyphenation pattern has been}%
\typeout{** loaded for the language `#1'. Using the pattern for}%
\typeout{** the default language instead.}%
\else
\language=\csname l@#1\endcsname
\fi
#2}}
\providecommand{\BIBdecl}{\relax}
\BIBdecl

\bibitem{open-domain-qa}
D.Moldovan, S.Harabagiu, and M.Pasca, ``The structure and performance of an
  open-domain question answering system,'' in \emph{Proceedings of the 38th
  Annual Meeting on Association for Computational Linguistics}, 2000, pp.
  563--570.

\bibitem{function-based-question-classification}
B.Fan, Z.Xingwei, and Y.Hau, ``Function based question classification for
  general qa,'' in \emph{Proceedings of Conference on Empirical Methods in
  Natural Language Processing}, 2010, pp. 1119--1128.

\bibitem{language-use}
S.K.Maity, A.Kharb, and A.Mukherjee, ``Language use matters: Analysis of the
  linguistic structure of question texts can characterize answerability in
  quora,'' \emph{CoRR}, 2017.

\bibitem{question-question}
C.~Shah, V.~Kitzie, and E.~Choi, ``Questioning the question -- addressing the
  answerability of questions in community question-answering,'' in \emph{47th
  Hawaii International Conference on System Sciences}.\hskip 1em plus 0.5em
  minus 0.4em\relax IEEE, Jan 2014, pp. 1386--1395.

\bibitem{not-answered-questions}
L.Yang, S.Bao, Q.Lin, X.Wu, D.Han, Z.Su, and Y.Yu, ``Analyzing and predicting
  not-answered questions in community-based question answering services,'' in
  \emph{AAAI'11}, 2011, pp. 1273--1278.

\bibitem{will-q-answered}
G.Dror, Y.Maarek, and I.Szpektor, ``Will my question be answered? predicting
  question answerability in community question-answering sites,'' in
  \emph{ECML- PKDD '13}, 2013.

\bibitem{bloomfilter}
B.~Bloom, ``Space/time trade-offs in hash coding with allowable errors,''
  \emph{Communications of the ACM}, vol.~13, no.~7, pp. 422--426, 1970.

\bibitem{lsh}
P.Indyk and R.Motwani, ``Approximate nearest neighbors: Towards removing the
  curse of dimensionality,'' in \emph{Proceedings of the 30th ACM Symposium on
  Theory of Computing}.\hskip 1em plus 0.5em minus 0.4em\relax ACM, 1998, pp.
  604--613.

\bibitem{plsh}
R.Panigrahy, ``Entropy based nearest neighbor search in high dimensions,'' in
  \emph{Proceedings of the 17th annual ACM-SIAM Symposium on Discrete
  Algorithms}.\hskip 1em plus 0.5em minus 0.4em\relax Society for Industrial
  and Applied Mathematics, 2006, pp. 1186--1195.

\bibitem{relu}
V.~Nair and G.~E. Hinton, ``Rectified linear units improve restricted boltzmann
  machines,'' in \emph{Proceedings of the 27th International Conference on
  International Conference on Machine Learning}, 2010, pp. 807--814.

\bibitem{adam}
D.~Kingma and J.Ba, ``Adam: A method for stochastic optimization,'' in
  \emph{Proceedings of the 3rd International Conference on Learning
  Representations (ICLR)}, 2015, pp. 1--13.

\bibitem{sqa-dataset}
P.~Rajpurkar, J.~Zhang, K.~Lopyrev, and P.~Liang, ``Squad: 100,000+ questions
  for machine comprehension of text,'' in \emph{Proceedings of the 2016
  Conference on Empirical Methods in Natural Language Processing}, 2016, pp.
  2383--2392.

\bibitem{kmeans++}
\BIBentryALTinterwordspacing
D.Arthur and S.Vassilvitskii, ``K-means++: The advantages of careful seeding,''
  in \emph{Proceedings of the Eighteenth Annual ACM-SIAM Symposium on Discrete
  Algorithms}, ser. SODA '07.\hskip 1em plus 0.5em minus 0.4em\relax
  Philadelphia, PA, USA: Society for Industrial and Applied Mathematics, 2007,
  pp. 1027--1035. [Online]. Available:
  \url{http://dl.acm.org/citation.cfm?id=1283383.1283494}
\BIBentrySTDinterwordspacing

\end{thebibliography}
\end{document}